\documentclass[10pt,twocolumn,letterpaper]{article}

\usepackage{iccv}
\usepackage{times}
\usepackage{epsfig}
\usepackage{graphicx}
\usepackage{amsmath}
\usepackage{amssymb}
\usepackage{multirow}
\usepackage{graphicx}
\usepackage[euler]{textgreek}
\usepackage{algorithm}
\usepackage{algpseudocode}
\usepackage{booktabs}

\usepackage[breaklinks=true,bookmarks=false]{hyperref}

\usepackage{pifont}

\newcommand{\sref}[1]{Sec.~\ref{#1}}
\newcommand{\eqqref}[1]{Eq.~\ref{#1}}
\newcommand{\figref}[1]{Fig.~\ref{#1}}
\newcommand{\figureref}[1]{Figure~\ref{#1}}
\newcommand{\tabref}[1]{Tab~\ref{#1}}
\newcommand{\tableref}[1]{Table~\ref{#1}}

\ificcvfinal\fi
\iccvfinalcopy
\begin{document}

\title{Improved Flood Insights: Diffusion-Based SAR to EO Image Translation}

\author{Minseok Seo, Youngtack Oh, Doyi Kim, Dongmin Kang, Yeji Choi\\
SI Analytics\\
70, Yuseong-daero 1689beon-gil, Yuseong-gu,
Daejeon, Republic of Korea\\
{\tt\small \{minseok.seo, ytoh96, doyikim, dmkang, yejichoi\}@si-analytics.ai}
}

\maketitle

\begin{abstract}
Driven by rapid climate change, the frequency and intensity of flood events are increasing.
Electro-Optical (EO) satellite imagery is commonly utilized for rapid response.
However, its utilities in flood situations are hampered by issues such as cloud cover and limitations during nighttime, making accurate assessment of damage challenging.
Several alternative flood detection techniques utilizing Synthetic Aperture Radar (SAR) data have been proposed.
Despite the advantages of SAR over EO in the aforementioned situations, SAR presents a distinct drawback: human analysts often struggle with data interpretation.
To tackle this issue, this paper introduces a novel framework, Diffusion-Based SAR to EO Image Translation (DSE).
The DSE framework converts SAR images into EO images, thereby enhancing the interpretability of flood insights for humans.
Experimental results on the Sen1Floods11 and SEN12-FLOOD datasets confirm that the DSE framework not only delivers enhanced visual information but also improves performance across all tested flood segmentation baselines.
\end{abstract}

\section{Introduction}
Under global warming conditions, the intensity and frequency of heavy precipitation and associated flooding events have increased in most regions \cite{allan2021ipcc, rohde2023floods}.
Flooding is one of the most prevalent natural disasters, and its damage leads to catastrophic consequences, especially in low-income countries \cite{rentschler2022flood}. 
 It is important and urgent to decide where to deploy the necessary resources to mitigate the damage and quickly recover from a crisis
%
The allocation of needed resources relies on precise information, which are collected through both manual and remote means.
%

Herein, Electro-Optical (EO) satellites have provided a broad and comprehensive view of the disaster-stricken region, surpassing the scope of on-site surveys by humans.
\begin{figure}[t!]
    \centering
    \includegraphics[width=1.0\columnwidth]{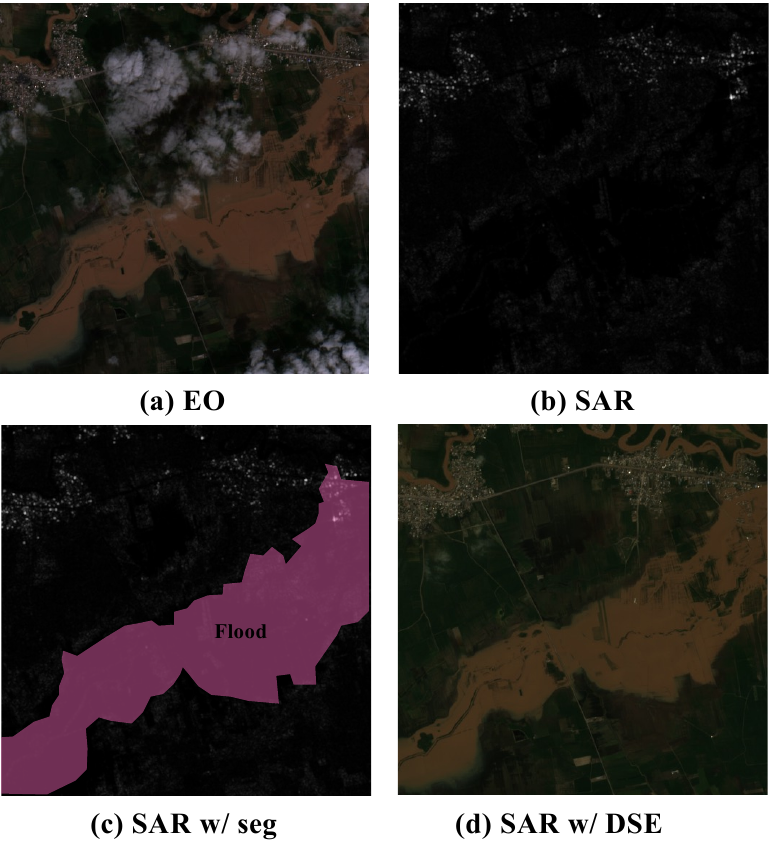}
    \caption{Examples of EO images and unprocessed SAR images from a flood-affected region. (a) depicts an EO image, (b) shows a SAR image, (c) presents the result of flood area segmentation using SAR imagery, and (d) displays a SynEO sample generated through the DSE framework.}
    \label{fig:motivation}
\end{figure}
One significant capability of EO satellites is the use of various optical channels to capture target images with high spatial resolution.

%
Satellite-based indexes, such as the Normalized Difference Water Index (NDWI) \cite{mcfeeters1996use} and Modified Normalized Difference Water Index (MNDWI) \cite{xu2006modification}, leverages variances in specific channels to monitor the water bodies and delineate the flood extent.
However, in EO satellite observations, cloud cover has obstructed the view of the region underneath, and often water and cloud shadows are misclassified (~\figref{fig:motivation}-(a)).  

It is because the sensors of EO satellites cannot penetrate dense cloud cover, and unfortunately, most flood events are due to heavy rains accompanied by thick clouds. Thus, the EO satellites are not suitable for real-time flood monitoring.
%
%
%
As an alternative, approaches that employ Synthetic Aperture Radar (SAR) data have been proposed \cite{tay2020rapid}.
SAR imaging holds the advantage of being unaffected by cloud cover and can operate under nighttime conditions, providing a flexible practice for disaster.

However, SAR images often contain impediments to interpretation, such as speckle noise, as shown in ~\figref{fig:motivation}-(b).
%
Hence, although a model appropriately estimates the inundated region as depicted in ~\figref{fig:motivation}-(c), people cannot easily affirm that model decisions are reliable without the aid of EO image (~\figref{fig:motivation}-(a)).
%
%
In response to this issue, we introduce the \textit{Diffusion-Based SAR to EO Image Translation} (DSE) framework, an innovative method to generate \textbf{\textit{clean}} Synthetic EO (SynEO) images from SAR inputs for enhanced flood monitoring and mapping.
Grounded in the Brownian Bridge Diffusion Model (BBDM)~\cite{Li2022BBDMIT}, our DSE framework stands as a robust image-to-image translation model based on the diffusion process.
Furthermore, we incorporate a self-supervised denoising method to enhance the clarity of generated images, thus improving the interpretability and usability of SAR images.
Concentrating on flood events, our work is designed to support decision-makers in their rapid and effective response to these natural disasters.
We leverage the DSE framework to generate EO-like images (SynEO) from SAR observations.
As demonstrated in ~\figref{fig:motivation}-(d), this enhanced view of the affected areas, which includes the SAR image and its corresponding SynEO, offers valuable insight for accurate flood monitoring.
%
%
To validate our DSE framework, we conducted assessments on the Sen1Floods11 \cite{bonafilia2020sen1floods11} dataset.
To confirm whether our SynEO truly aids SAR experts, we conducted experiments using the SEN12-FLOOD \cite{rambour2020sen12} dataset.
Through our comprehensive analysis, we found that compared to cloud-free EO images, our SynEO only exhibited a negligible performance drop of approximately 1\%.


\begin{figure*}[t!]
    \centering
    \includegraphics[width=1.8\columnwidth]{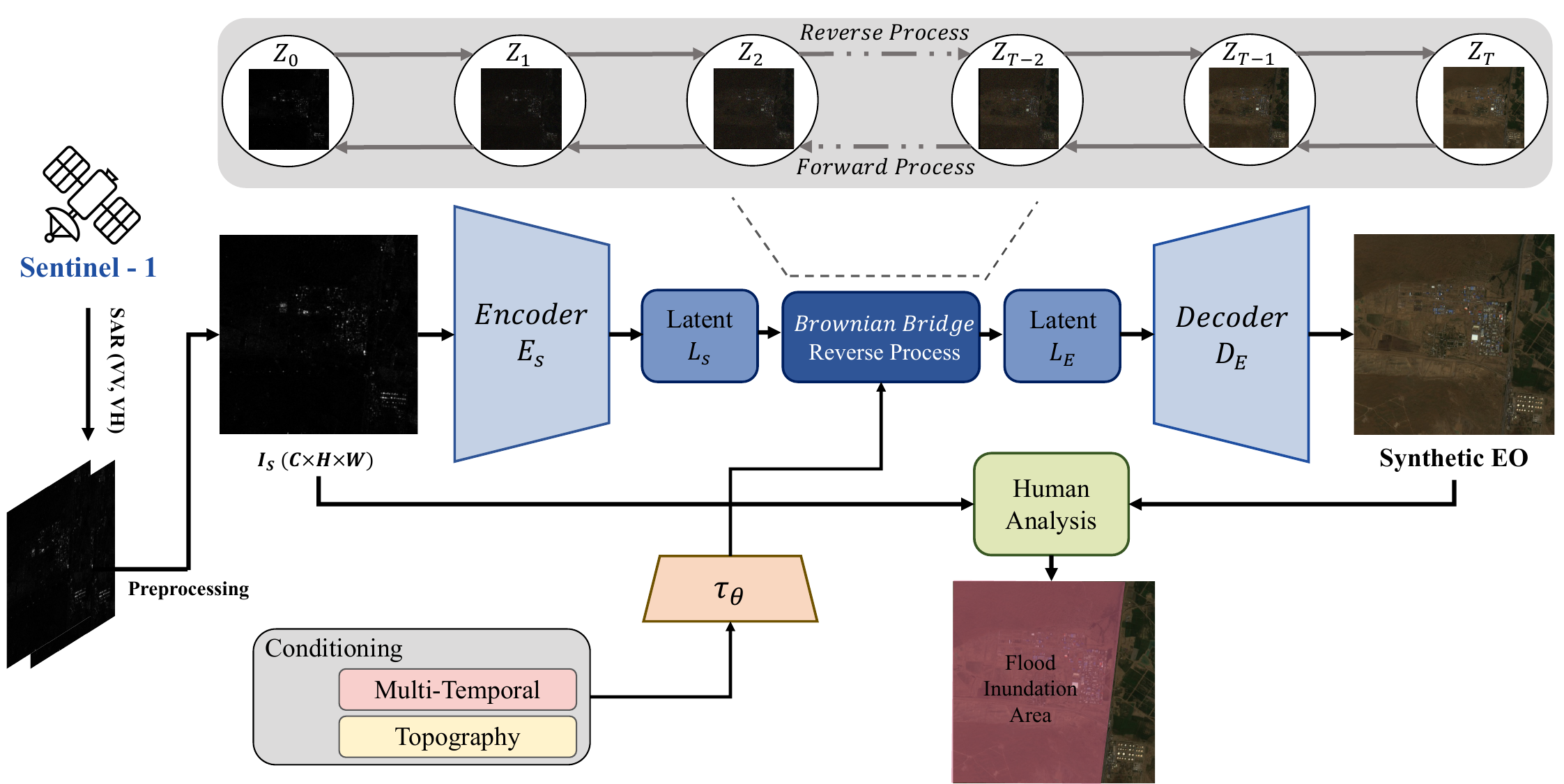}
    \caption{Overview of the DSE framework. The DSE framework takes in the SAR image, applies a self-supervised denoising method, and then carries out diffusion-based SAR2EO image translation. Subsequently, the generated EO and corresponding SAR images are reviewed by the analysts for the purpose of flood mapping.}
    \label{fig:method}
\end{figure*}
\section{Related Work}

\subsection{EO Imagery for Disaster Management}
Earth observation through satellite delivered a major breakthrough in disaster risk management ~\cite{LeCozannet2020SpaceBasedEO}.
With the EO satellite images, various types of disasters and hazards could be monitored and managed.
For example, inland flooding \cite{Lei2022FloodDM,kordelas_automatic_2019,Kordelas2018FastAA}, subsidence \cite{Cigna2021SatelliteIS}, landslides \cite{Ghorbanzadeh2022LandslideDU}, earthquakes and volcanic eruptions \cite{Schneider2020SatelliteOO}, or wildfires \cite{Chen2022CaliforniaWS}.

Among those disasters, flood is the single most hydrometeorological hazard causing substantial losses. \cite{bonafilia2020sen1floods11}
For rapid and efficient monitoring of flooding by EO image, there has been a wide range of methods proposed.
Thesholding NDVI (Normalized Difference Vegetation Index) or NDWI (Normalized Difference Water Index) based models had been simple yet effective \cite{Kordelas2018FastAA}.
Hence they are still utilized as a reference for models or human analysts \cite{bonafilia2020sen1floods11}.
Recently, machine learning based supervised classification algorithms have been used, including Random Forests \cite{Ko2015ClassificationOP}, decision trees \cite{Acharya2016IdentificationOW}, support vector machine (SVM) \cite{Sarp2017WaterBE} and the perceptron model \cite{Mishra2015AutomaticEO}, neural networks \cite{Skakun2012ANN}.
Those classification-based approaches achieve relatively higher accuracy than index thresholding methods while it requires ground truth data to select appropriate training samples.

The major drawback of using EO images for disaster management is that it cannot guarantee consecutive monitoring when the Area of Interest (AoI) is covered by clouds or is in night time. \cite{Nemni2020FullyCN}

\subsection{SAR Imagery for Disaster Management}
Compared to the EO-based system, observation through radar is another important approach that accelerated the era of earth observation and disaster management.
SAR provides a stable monitoring environment even in the case of cloud-covered or night-time situations. \cite{Kordelas2018FastAA}

In a similar sense to EO imagery, SAR has been used to monitor various types of disasters (e.g., flood \cite{Nemni2020FullyCN}, landslide \cite{Florentino2016ImplementationOA}, tsunami \cite{endo2018new}, etc.).
Especially, SAR has shown noticeable advantages against EO imagery in studying water bodies \cite{Ji2018EarthquakeTsunamiDL}.
Widely used water index in EO data has suffered one drawback: bands associated with the near-infrared and short-saved infrared can present a loss of resolution compared to the RGB ones.

With SAR imagery, there have been several well-known methods for flood monitoring.
Conventional method, similar with EO imagery, is threshold based method \cite{White2015ACO}.
Recently, machine learning based algorithms have been proposed and shown comparable or better performance than threshold based method in terms of generalization \cite{ban2020near}.

\subsection{SAR to EO Image Translation}
Despite of high performance of those methods, SAR based model has an intrinsic weakness; the interpretation of the SAR images is extremely difficult because of speckle noise and innately different data representation than EO image.
Inevitably, these interpretation difficulties require annotated data \cite{Wang2019SARtoOpticalIT}.
Recent automated disaster monitoring models are based on neural networks, which means they are heavily data-driven. \cite{nemni2020fully, seo2023self, noh2022unsupervised}
Therefore, the scarcity of data is disrupting the potential of SAR in many applications.

To alleviate this disadvantage of SAR datasets, visualization techniques for SAR imagery have been studied.
Most direct and simple methods are re-normalization methods such as histogram equalization \cite{Li2014AutomaticSI} and thresholding-based methods \cite{Li2011AnAM}.
 
Although it helps to recognize the overall structure of the scene, still detailed features are not easily acquired.
One straightforward solution is denoising the speckle noise \cite{Deledalle2015NLSARAU} while it shows its effectiveness in limited environments.
Hence, recent studies proposed SAR to EO image translation method in the sense that EO imagery is most straightforward for humans~\cite{low2023multi}.

Generative adversarial Networks (GAN) based Image translation has been proposed \cite{Wang2019SARtoOpticalIT,Reyes2019SARtoOpticalIT}.
However, GAN-based models frequently suffer from mode collapse problems that significantly degrade the quality of generated data.
This is the first application of flood disaster monitoring to our best knowledge.
We propose DSE framework which does not suffer from those issues.

\subsection{Diffusion Models}
For the new scheme of generative model, diffusion models has been recently highlighted. \cite{ho2020denoising}
Diffusion models generate data by sequantial denoising steps.
Initially, diffusion models were treated as Markov chain with Gaussian noise and recently we interpret it as a stochastic differential equation (SDE).
A remarkable property of this SDE is the existence of an ordinary differential equation (ODE), dubbed the Probability Flow (PF) ODE by ~\cite{Song_Sohl-Dickstein_Kingma_Kumar_Ermon_Poole_2021}.

With the great success in image generation quality, there are recent trials to use diffusion model as conditional generation model \cite{wang2018high}.
However, those trial succeeded in limited application.
To handle with the limitation of diffusion models in image translation task, \cite{Li2022BBDMIT} exploited the formulation of Brownian bridge to model the stochastic process of image translation in latent space.

In satellite imagery domain, also, diffusion models have been broadly applied \cite{perera2023sar}.
Our proposed method also get advantage of diffusion model's generation power to make SAR more interpretable.
For the detailed description of our proposed method, please refer to the following sections. 
\section{Method}
In this section, BBDM \cite{Li2022BBDMIT}, which is the basis of the DSE framework, is first briefly described, and then the preprocessing, model, and function are sequentially explained in detail.
Please note that the reverse process of DSE aligns perfectly with that of BBDM, so we won't delve into the details of the reverse process in this paper.

\subsection{Brownian Bridge Diffusion Model (BBDM)}
Given two datasets, $X_{A}$ and $X_{B}$, originating from domains $A$ and $B$ respectively, the purpose of image-to-image translation is to ascertain a function that establishes a mapping from domain $A$ to domain $B$.
While numerous image-to-image translation methods based on conditional diffusion models have been proposed, they are not intuitively suited for the task as its translation process seamlessly converts a noise back into an image, not image to image.
Moreover they does not have a clear theoretical guarantee because of their complex conditioning algorithm based on attention mechanism.
BBDM, however, provides a method for image-to-image translation grounded in the Brownian diffusion process which avoid leveraging complex conditioning algorithm.

Reffering to the original BBDM, we also conduct the process in the latent space of VQGAN\cite{esser2021taming}.
Following the convention, let $(x,y)$ denote the paired training data from $X_A$ and $X_B$, each.
For simplicity, we use $x$ and $y$ to denote the corresponding latent features ($x := L_S(x), y:= L_E(y)$).
The forward diffusion process of Brownian Bridge is defined as:
\begin{align}
q_{BB}(x_{t}|x_{0},y) & = \mathcal{N}(x_{t};(1-m_{t})x_{0}+m_{t}y, \delta_{t}I) , \label{eq:eq11}
\\
\ x_{0} & = x, m_{t} = \frac{t}{T}
\label{eq:eq1}
\end{align}
where $T$ is the total steps of the diffusion process, $\delta_{t}$ is the
variance.

The forward diffusion of the Brownian Bridge process provides only the marginal distribution at each time step $t$, as shown by the transition probability in ~\eqref{eq:eq11}.
However, for training and inference, it is essential to deduce the forward transition probability $q_{\text{BB}}(x_{t}|x_{t-1}, y)$. 
In the original BBDM, given an initial state $x_{0}$ and a destination state $y$, the intermediate state $x_{t}$ can be computed in discrete form as follows:
\begin{align}
x_{t} & = (1-m_{t})x_{0}+m_{t}y+\sqrt{\delta_{t}}\epsilon_{t},
\\
x_{t-1} & = (1-m_{t-1})x_{0}+m_{t}y+\sqrt{\delta_{t-1}}\epsilon_{t-1}
\label{eq:eq2}
\end{align}
here, $\epsilon_{t}$, $\epsilon_{t-1} \sim \mathcal{N}(0, I)$. 

However, in the SAR2EO task, diversity isn't as crucial as in the original BBDM.
Rather, the emphasis is on \textbf{\textit{prediction}} that closely aligns with the actual outcome.
For instance, in the SAR2EO task, the goal is to generate images that are akin to the actual EO image or resemble the distribution of training EO images, instead of producing a variety of colors and textures like \textbf{\textit{generation}}.
Consequently, we sample $\epsilon$ from the target distribution rather than the standard normal distribution $\mathcal{N}(0, I)$. Moreover, in the reverse process, we set the size of $\epsilon$ to $ \epsilon \times 0.1$. This adjustment brings the SAR2EO task closer to prediction.
\subsection{Pre-processing}
\label{sec:deno}
SAR images are intrinsically speckled due to the way they are generated, which can be captured by the following mathematical model for multiplicative speckle noise $N$:
\begin{equation}
Y = XN,
\label{eq:s}
\end{equation}
where $Y$ is the observed SAR intensity, $X$ is the speckle-free or clean image, and $N$ is the speckle noise.

Generally, it is postulated that $N$ conforms to a Gamma distribution, characterized by a mean of 1 and a variance of $1/L$, where $L$ represents the number of `looks' in the multi-look process. The probability density function of this particular distribution can be formulated as follows:

\begin{equation}
p(N) = \frac{1}{\Gamma(N)}L^{N}N^{L-1}e^{-LN},
\label{eq:speckle_noise_model}
\end{equation}

where $\Gamma(.)$ is the Gamma function. This formulation gives a more comprehensive account of the characteristics of speckle noise in SAR images.
As the DSE framework applies a diffusion-based image-to-image translation model, it is designed to predict both the SAR image and the noise added during the forward process.

\begin{figure}[t!]
    \centering
    \includegraphics[width=\columnwidth]{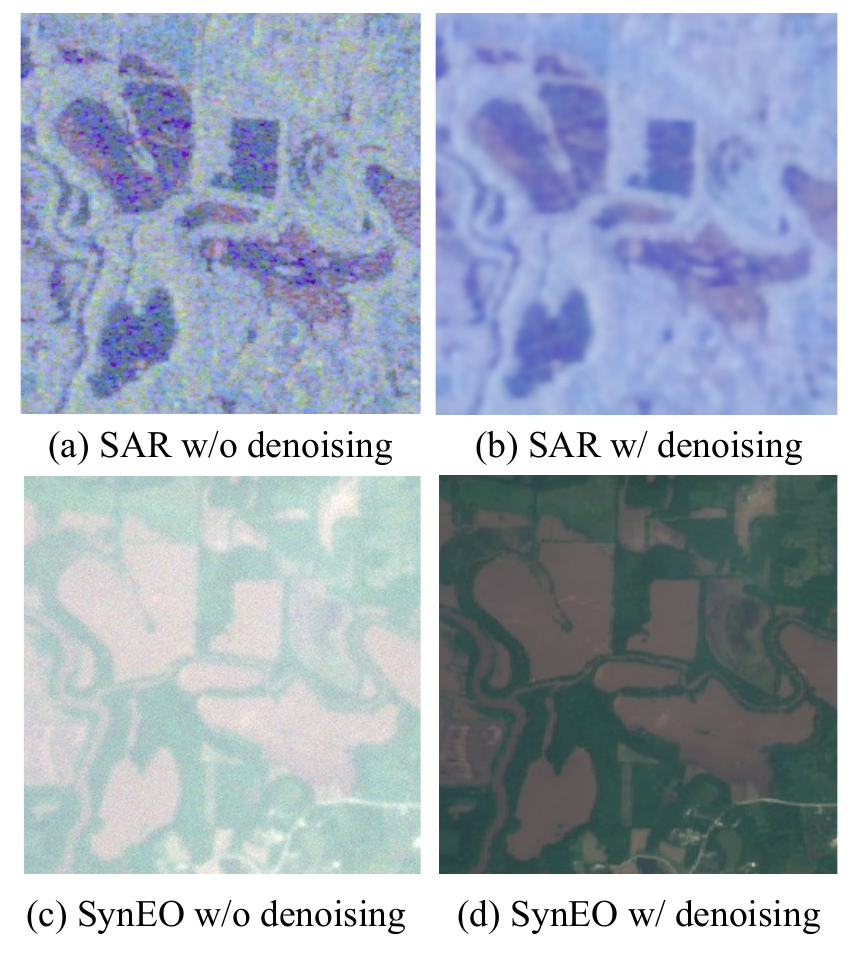}
    \caption{Qualitative comparison between original SAR images and SAR images denoised using the self-supervised method.}
    \label{fig:deno}
\end{figure}

However, the model may struggle to distinguish between the speckle noise inherent to SAR imaging, as depicted by ~\eqqref{eq:speckle_noise_model}, and additional noise introduced during the forward process.
Consequently, remnants of noise may still be present after the DSE framework is applied, as shown in ~\figref{fig:deno}-(b).

To address this issue, we de-noise the SAR image using a blind-spot based self-supervised denoising method beforehand we leverage them in the image-to-image translation process.
The conventional blind-spot based approaches inherently assume that the noise is independent of the clean image~\cite{lehtinen2018noise2noise}, a condition not fulfilled by SAR images produced via ~\eqqref{eq:speckle_noise_model}.
To mitigate this, we applied the ~\cite{zhang2023mm} method, a variant of blind-spot techniques utilizing diverse kernels.
~\figref{fig:deno}-(b,d) presents the results of de-noising using the ~\cite{zhang2023mm} method and the SAR2EO generation results using the de-noised SAR image with the DSE framework.
\begin{table*}[t!]
\centering
\resizebox{1.7\columnwidth}{!}{%
\begin{tabular}{c|c|ccccc}
\hline \hline
\multirow{2}{*}{\textbf{Method}} & \multirow{2}{*}{\textbf{Modality}} & \multicolumn{5}{c}{\textbf{Metric}}                         \\ \cline{3-7} 
                        &                           & Pre.(0/1) & Rec. (0/1) & F1 (0/1) & IoU (0/1) & ACC \\ \hline
U-Net~\cite{ronneberger2015u} (ResNet34)        & SAR                       & 0.9622/0.4267         & 0.9796/0.2871           & 0.9708/0.3409         & 0.9433/0.2146          & 0.9442    \\
U-Net~\cite{ronneberger2015u} (ResNet34)        & EO                        & 0.9930/\textbf{0.7833}         & \textbf{0.9858}/0.8642            & \textbf{0.9894}/\textbf{0.8162}        & \textbf{0.9790}/\textbf{0.6928}          & \textbf{0.9800}    \\
U-Net~\cite{ronneberger2015u} (ResNet34)        & SAR+EO                    & 0.9894/0.7673         & 0.9840/0.7954           & 0.9867/0.7685         & 0.9455/0.6291          & 0.9749    \\ \hline
U-Net~\cite{ronneberger2015u} (ResNet34)        & SAR+SynEO                 & 0.9862/0.7085         & 0.9817/0.7352           & 0.9838/0.7041         & 0.9683/0.5556          & 0.9695    \\
U-Net~\cite{ronneberger2015u} (ResNet34)        & SynEO              & \textbf{0.9931}\slash0.4697         & 0.9440\slash\textbf{0.8723}           & 0.9678\slash0.6070        & 0.9378\slash0.4377          & 0.9408   \\ \hline \hline
\end{tabular}%
}
\label{tab:sen12seg}
\caption{Comparison of precision, recall, F1-Score, IoU, and accuracy of flood segmentation results for different modalities for the SEN12-FLOOD Dataset. In metric, 0 is the performance when detecting a background, and 1 is the performance when detecting a flood. Also, in modalities, the + sign means that each image is concatenated to a channel axis.}
\end{table*}
\begin{table}[h!]
\centering
\resizebox{0.7\columnwidth}{!}{%
\begin{tabular}{c|ccc}
\hline \hline
\textbf{Method}    & \textbf{PSNR} & \textbf{SSIM} & \textbf{LPIPS} \\ \hline
Pix2PixHD~\cite{wang2018high}&   31.09   &   0.81    &  0.116   \\
BBDM~\cite{Li2022BBDMIT}      &   29.20   &  0.74    & 0.124    \\ \hline
DSE       &   32.43   &   0.84   &    0.109 \\ 
DSE+multi-temporal       &  \textbf{34.94}    &   \textbf{0.87}   & \textbf{0.082}    \\ \hline \hline
\end{tabular}%
}
\caption{Comparison results of the DSE framework with the commonly employed SAR2EO baselines, pix2pixHD, using a test set derived from the SEN12-FLOOD dataset, where missing or cloud-affected data points have been excluded.}
\label{tab:sen12}
\end{table}
\subsection{SAR2EO Image Translation}
~\figureref{fig:method} provides a schematic overview of the DSE framework.
DSE accepts VV and VH channels, transforming them into a 3-channel image in the form of (VV, VH, (VV+VH)/2.).
Subsequently, the image undergoes self-supervised denoising via the method described in ~\sref{sec:deno} to minimize speckle noise.
This processed image is then fed into the diffusion model-based SAR2EO translation network to generate a synthetic EO (SynEO) image.
Ultimately, through this process, the synthetic EO (SynEO) image, paired with the corresponding SAR image, is presented to the SAR experts.
This combined imagery provides additional insights into flood conditions, aiding the evaluator in making a more informed assessment of the flood extent.
\section{Experiments}
In this section, we conducted experiments on two flood datasets, Sen1Floods11~\cite{bonafilia2020sen1floods11} and SEN12-FLO~\cite{w6xz-s898-20}, to evaluate the effectiveness of the DSE framework. 
The Sen1Floods11 dataset is characterized by a high presence of clouds, while the SEN12-FLOOD dataset, a multi-temporal dataset, has fewer clouds. 
It's important to note that for both datasets, we conducted experiments after removing clouds using the QA60 method~\cite{yu2023sar2eo}.

\paragraph{Training \& Test Datasets} 
The Sen1Floods11 dataset we used for SAR2EO is a dataset consisting of Sentinel-1 and Sentinel-2 flood event imagery sourced from Google Earth Engine \cite{bonafilia2020sen1floods11}.
The reference flood maps were generated applying specific thresholds.
The data was segmented into $512 \times 512$ pixel chips, with a select 446 chips manually labelled for validation.
The remainder of the data was randomly split into a 60-20-20 distribution for training, validation, and testing,
incorporating non-hand-labeled Sentinel-1 and Sentinel-2 data for weakly supervised training. Note that our method does not generate clouds, so the test set was selected with cloud-free images of the same volume.

The SEN12-FLOOD Dataset, which we used for our semantic segmentation and 
qualitative comparison, is compiled from 336 time series featuring Sentinel 1 and Sentinel 2 images of regions that experienced significant flooding during the winter of 2019.
The data collection period spans from December 2018 to May 2019, with the observed areas primarily located in East Africa, South West Africa, the Middle-East, and Australia.
A sequence corresponds to multiple 512x512 tiles, each representing a crop from a specific acquisition
\paragraph{Evaluation metric}
In our flood segmentation task, we evaluated model performance through F1, precision, recall, and IoU metrics, while the image-to-image translation task was assessed via PSNR, SSIM, and LPIPS metrics.
Note that due to the lack of color information in SAR images, PSNR may not provide fully meaningful insights, thus necessitating the addition of the SSIM and LPIPS metric.
Even though the SAR2EO task is a predictive task with definite GT paired with input, low PSNR or SSIM does not indicate the model's output is simply wrong.
As SAR data aggregated significantly different wavelength range compared to that of EO, there can be multiple (diverse) probable EO images corresponding to one SAR image and vice versa.
It leads to the necessity of utilizing LPIPS as a supplementary indicator.
%
\begin{figure*}[t!]
    \centering
    \includegraphics[width=2.0\columnwidth]{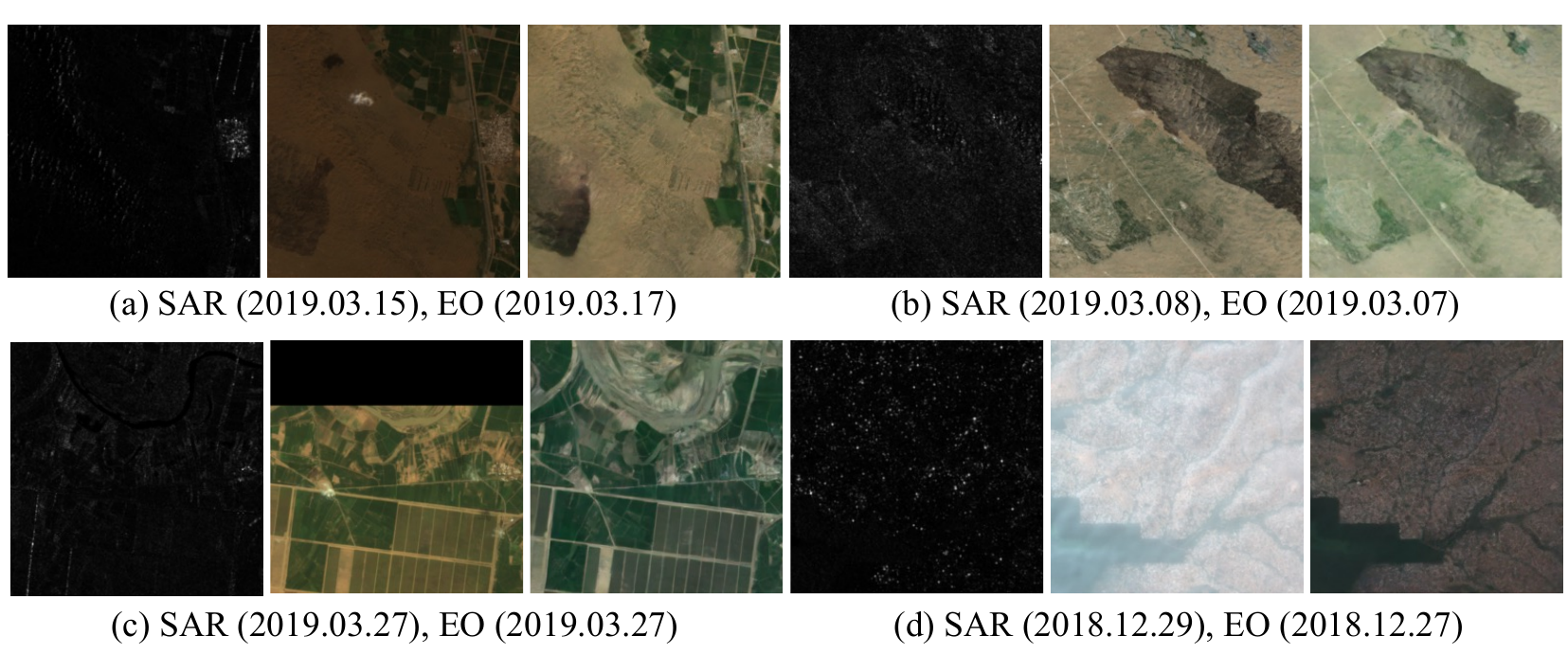}
    \caption{Results of applying the DSE framework on the SEN12-FLOOD dataset. From left to right: SAR, EO, and Synthetic EO. (Note that SAR image is (VV, VH, (VV+VH)/2.) 3-channel image, but it is visualized as a gray scale.)}
    \label{fig:vis}
\end{figure*}
\paragraph{Implementation details}
The DSE framework comprises two main components: a pretrained VQGAN~\cite{esser2021taming} model and our newly proposed Diffusion-Based SAR to EO model. VQGAN typically excels at reducing both computational load and inference time.
In the interest of constructing an efficient diffusion model, we adopted the VQGAN model, as featured in the Latent Diffusion Model~\cite{rombach2022high}.
We set the number of time steps for the Brownian Bridge to be 1000 during the training stage and employed 200 sampling steps during the inference stage, thus striking a balance between sample quality and computational efficiency. It's important to note that our baseline code builds upon BBDM~\footnote{\url{github.com/xuekt98/BBDM}}, and all hyperparameters not explicitly mentioned align with those used in BBDM~\cite{Li2022BBDMIT}.

In addition, the semantic segmentation model we employed in our experiment was sourced from repository~\footnote{\url{github.com/qubvel/segmentation_models.pytorch}}, and we utilized a fundamental U-Net based on ResNet34. All hyperparameters strictly adhere to the default settings.
\subsection{Quantitative Results}
\begin{figure*}[t!]
    \centering
    \includegraphics[width=2.0\columnwidth]{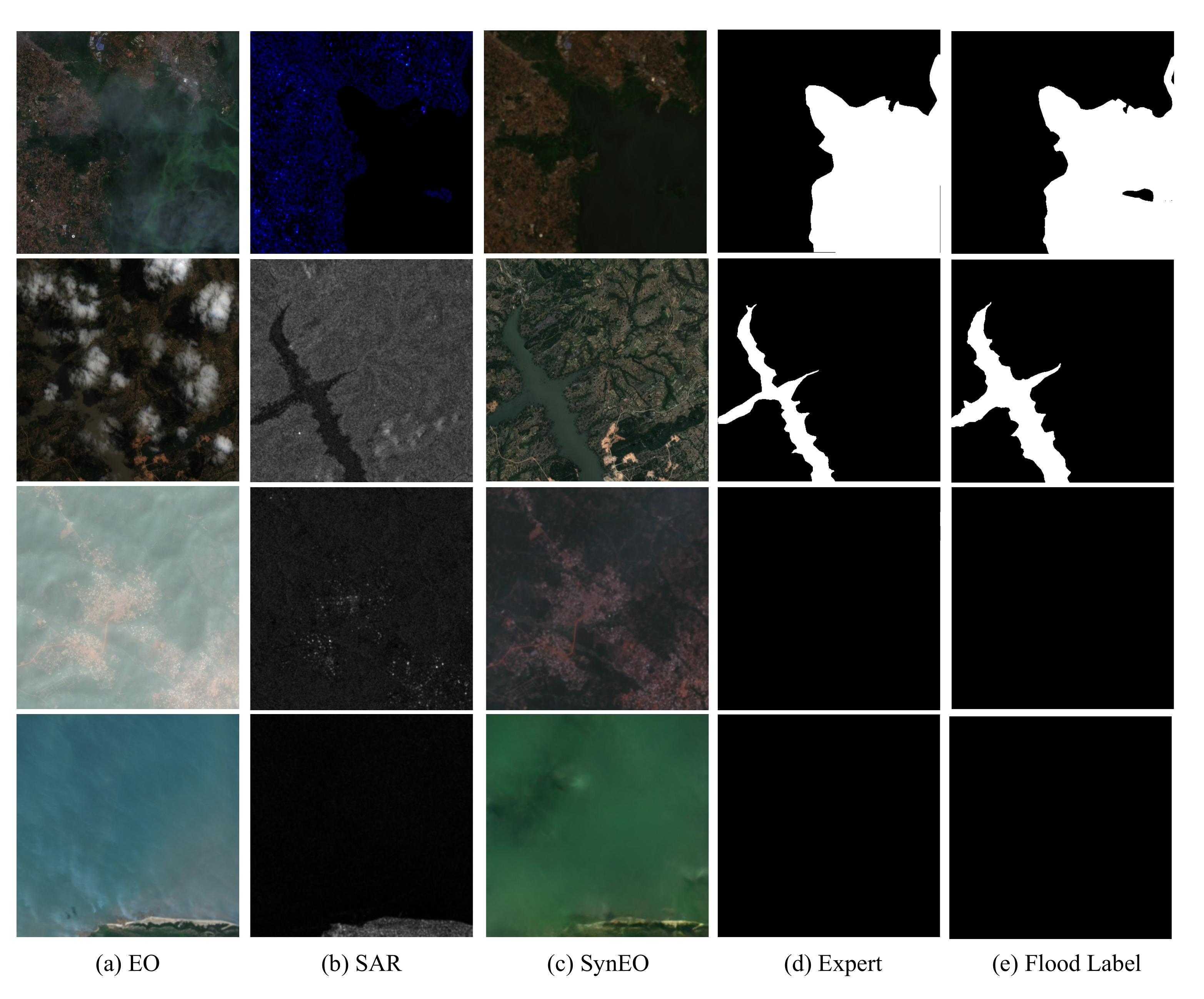}
    \caption{Comparison of flood detection by SAR experts using pairs of EO and SAR images versus pairs of SynEO and SAR images. Please note that we provided SAR experts with both three-channel SAR images (VV, VH, (VV+VH)/2) and one-channel images simultaneously. The SAR images included in the figure have been identified by SAR experts as being more conducive to their analyses.}
    \label{fig:vis}
\end{figure*}
\paragraph{Image-to-Image Translation}
~\tableref{tab:sen12} provides the experimental findings from the SEN12-FLOOD dataset's test subset, with clouds and data voids excluded, showcasing the proficiency of the DSE framework in synthesizing SAR2EO imagery.
It's crucial to note that the temporal alignment between the multi-temporal SAR and EO data within the SEN12-FLOOD dataset isn't precise.
Accordingly, we've matched the EO data from the nearest date to the reference SAR imagery.

As displayed in ~\tabref{tab:sen12}, among the compared baselines, DSE demonstrated superior performance in terms of PSNR, SSIM, and LPIPS.
Furthermore, the most favorable performance was recorded when multi-temporal SAR (spatially registered and temporally random) was utilized as an input.
These results confirm that the DSE framework operates efficiently under these conditions.
Regrettably, our experimented dataset did not provide topographic information, thus we were unable to test the model under this scenario.
Future work aims to improve model accuracy by introducing topographic information as an additional condition.
\paragraph{Flood Segmentation}
~\tabref{tab:sen12seg} displays the quantitative comparison of flood segmentation results utilizing SAR, EO, and SynEO.
As indicated in the table, the flood segmentation solely based on SAR exhibits the lowest performance in terms of Recall, Rec., F1 score and IoU.
This is primarily due to the irregular pattern for the flood area in our single-temporal SAR experimental setting, as opposed to a multi-temporal SAR setting as in these referenced works~\cite{mateo2021towards}.
It's worth noting that even if the intensities are identical in two different SAR images, they could correspond to distinct objects.

The best result was achieved through segmentation using solely EO images.
This is because, for the sake of a fair experiment, we employed only cloud-free images in the test set; hence, the EO image appears the clearest and is the closest representation of the correct answer. 

Although the performance was lower than the EO image for the cloudless test set, the IoU when using SAR and SynEO was 0.555, which was almost 30 higher than when only SAR was used. Note that EO video is unavailable at night and in cloudy conditions.
These experimental results suggest that the SynEO can provide a level of information to the segmentation model that is comparable to clear, cloud-free EO images.
%
%
%
%
\paragraph{Human analysis}
\begin{table}[h!]
\centering
\resizebox{0.5\columnwidth}{!}{%
\begin{tabular}{c|c}
\hline \hline
\textbf{Modality}    & \textbf{IoU}  \\ \hline
SAR + EO      &   \textbf{0.5532}       \\
SAR + SynEO       &   0.5464 ($\downarrow 0.68$)   \\ \hline\hline
\end{tabular}%
}
\caption{Comparison of flood detection by SAR experts using pairs of EO and SAR images versus pairs of SynEO and SAR images. Note that this value represents the average result of flood mapping performed by five SAR experts.}
\label{tab:tab2}
\end{table}
The outcomes presented in ~\tabref{tab:tab2} represent flood mapping results derived by a SAR expert utilizing (EO,SAR) and (SynEO,SAR) pairs.
As evident from the table, simple EO images yield relatively low performance as they predominantly rely on SAR data due to obstructions like clouds, cloud shadows, and fog.
In contrast, SynEO images display superior performance compared to (EO, SAR) pairs as they offer additional information to the examiner devoid of obstructions such as clouds, cloud shadows, and fog.

Our experimental outcomes demonstrate that SynEO images can effectively be used as auxiliary data when EO images are impaired by conditions such as fog, night, and clouds.
Furthermore, the potential utility of SynEO data in circumstances where matched EO data cannot be secured is suggested.
It should be highlighted, however, that in our application, SynEO images may not prove particularly beneficial in situations where both clean SAR and corresponding EO images are available.
Given that our SynEO is based on SAR, it may incorporate errors.
Hence, in our application setting, SynEO was not used in isolation, but rather served as a supplemental resource for SAR.
\subsection{Qualitative Results}
\paragraph{Image-to-Image Translation}
~\figureref{fig:vis} showcases the utilization of the DSE framework on the SEN12-FLOOD dataset.
In the SEN12-FLOOD dataset, SAR and EO images are not always aligned by date, so the closest EO images were selected for both training and inference.
As demonstrated in the figure, the DSE framework is capable of generating synthetic EO images with remarkable clarity. 
Specifically, ~\figref{fig:vis}-(a) reveals that the flood area is accurately represented, and notably, without the formation of clouds or cloud shadows.
Further, as indicated in ~\figref{fig:vis}-(c), while the original EO image suffers from data loss, hence the absence of the top portion of the image, the synthetic EO (SynEO) images derived from SAR data are not afflicted by such data loss.
Finally, as depicted in ~\figref{fig:vis}-(d), in contrast to the original EO image, the generated image demonstrates the effective removal of haze.

\begin{figure}[t!]
    \centering
    \includegraphics[width=\columnwidth]{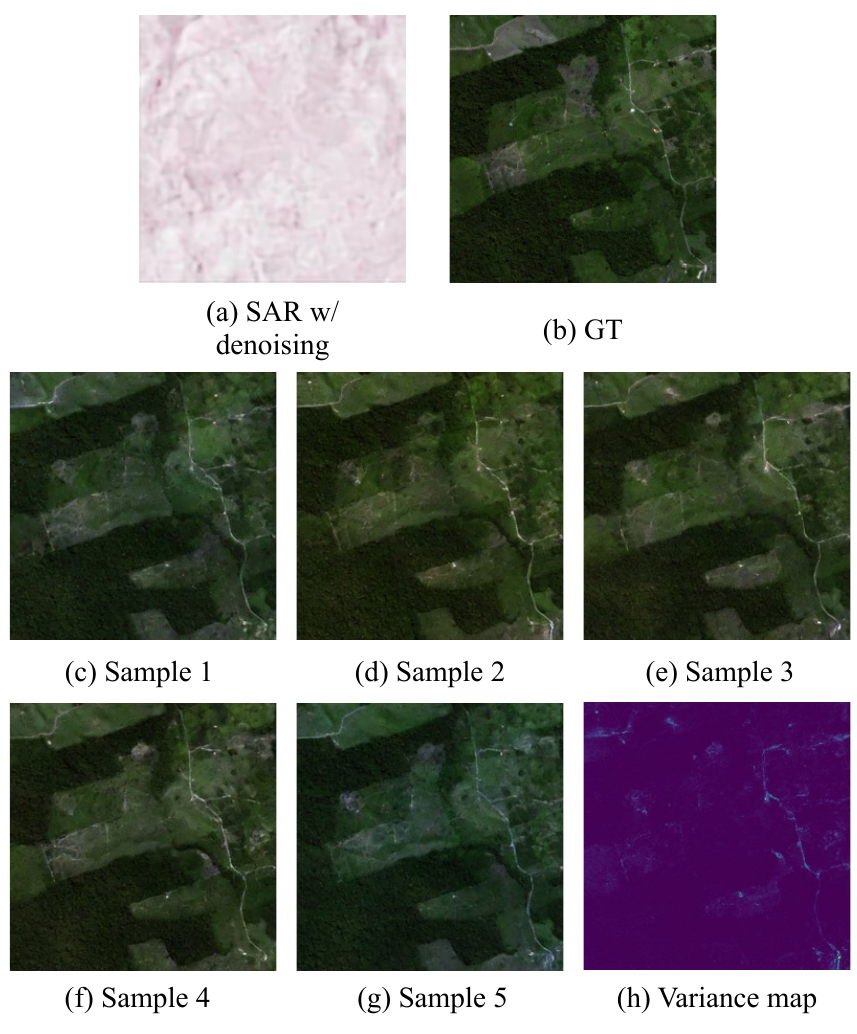}
    \caption{Various output samples generated by the DSE framework. The diffusion-based image-to-image translation allows for the creation of multiple samples, enabling uncertainty measurement through ensemble methods.}
    \label{fig:vari}
\end{figure}

\paragraph{Human analysis}
~\figureref{fig:vis} provides a qualitative comparison of flood mapping results obtained from (SynEO and SAR) and (EO, SAR) images, as analyzed by SAR experts.
As depicted in the figure, the use of EO images becomes challenging in conditions such as clouds and fog.
Furthermore, deriving flood maps from simple SAR images is complex, due to the inherent nature of these images.
Specifically, SAR images may display dark areas over water bodies or extremely flat surfaces where radar signals are reflected away from the sensor, and in regions of dense vegetation where signals are scattered in various directions.
As depicted in the figure, SynEO can support SAR experts under these conditions by supplementing the SAR image with additional information.
\paragraph{Variance map}
Differentiating from existing models, the DSE framework emerges as the pioneering diffusion model-based approach to SAR2EO.
This feature allows it to generate multiple varied samples. In practical terms, the objective of the SAR2EO task is to create real-time SynEO images that align closely with SAR images.
However, given that SynEO images derive from SAR data rather than from actual EO, it presents challenges in determining which sections to trust during expert analysis.
As demonstrated in ~\figref{fig:vari}, the DSE's capacity to produce an array of samples and a corresponding variance map, as seen in ~\figref{fig:vari}-(d), provides a unique advantage.
This benefit allows interpreters to place confidence in areas of consistency across the samples, focusing more on the SAR over the SynEO in areas of inconsistency. We firmly believe that such features embody considerable promise for the continued evolution of SAR2EO.
\section{Limitation and Future Work}
\paragraph{Limitation} In this study, we proposed a DSE framework designed to enhance the human interpretability of SAR.
However, the DSE framework requires the use of both spatial (strict) and temporal (near-time) pairs of SAR and EO for image-to-image translation.
Despite not necessitating flood labeling, obtaining and registering paired images can often prove challenging.
Consequently, this can result in significant costs, which constitutes a notable limitation of our current work.

\paragraph{Future work}
In our future work, we plan to create a more cost-effective DSE framework by effectively applying unpaired image-to-image translation from SAR to EO. Additionally, we aim to broaden its applicability to a diverse range of disaster scenarios, including earthquakes and forest fires, in addition to floods.

\section{Conclusion}
In this paper, we introduced a Diffusion-Based SAR to EO Image Translation (DSE) framework, aiming to improve human analysis of floods.
The DSE was designed to address two central issues.
First, exploiting EO imagery for flood mapping frequently suffers from impracticability in cloud cover or night-time while it is simply interpretable so that handy for labeling and analysis.
Second, the alternative deep learning-based SAR flood detection methodology, which was proposed to remedy this problem, demands a substantial volume of labeled flood data.
%
In order to exploit advantages of EO and SAR simultaneously, we proposed the DSE framework, a SAR-to-EO translation scheme which effectively contributes to human analysis of flood.
%
%
We validated our algorithm on the Sen1Floods11 and SEN12FLO datasets and obtained quantitative and qualitatively significant results.
We hope that our research will be widely used in flood detection.

{\small
\bibliographystyle{ieee_fullname}
\bibliography{egbib}
}

\end{document}